\documentclass[runningheads]{llncs}
\usepackage[T1]{fontenc}
\usepackage{graphicx}
\usepackage{caption}
\usepackage{subcaption}
\usepackage{amsmath}
\usepackage{bbm}
\usepackage{commath}
\usepackage{xcolor}
\begin{document}
\title{Semi-supervised Contrastive Regression for Estimation of Eye Gaze}
%
%
\author{Somsukla Maiti\inst{1} \and
	Akshansh Gupta\inst{1}}
\authorrunning{Maiti et al.}
\institute{CSIR- Central Electronics Engineering Research Institute, Pilani 
	\email{somsuklamaiti@gmail.com, somsukla@ceeri.res.in; akshanshgupta@ceeri.res.in}\\}

\maketitle              
\begin{abstract}
With the escalated demand of human-machine interfaces for intelligent systems, development of gaze controlled system have become a necessity. Gaze, being the non-intrusive form of human interaction, is one of the best suited approach. Appearance based deep learning models are the most widely used for gaze estimation. But the performance of these models is entirely influenced by the size of labeled gaze dataset and in effect affects generalization in performance. 

This paper aims to develop a semi-supervised contrastive learning framework for estimation of gaze direction. With a small labeled gaze dataset, the framework is able to find a generalized solution even for unseen face images. In this paper, we have proposed a new contrastive loss paradigm that maximizes the similarity agreement between similar images and at the same time reduces the redundancy in embedding representations. Our contrastive regression framework shows good performance in comparison to several state of the art contrastive learning techniques used for gaze estimation. 

\keywords{Gaze Estimation \and Contrastive Regression \and Semi-supervised learning \and Dilated Convolution}
\end{abstract}

\section{Introduction}
Eye gaze is one of the most widely used human mode for developing human-machine interface system. Gaze controlled interface has become quite popular for virtual reality (VR) applications \cite{konrad2020gaze}, navigation control of ground and aerial robots \cite{gerber2020self}, control of robotic arms for surgery \cite{ferrier2022measuring} and other commercial applications. With the advances in deep learning methods, gaze tracking has become an achievable task.

Appearance-based models are the most prominent approach that does not require an expensive and subject-dependent eye modeling. It mainly relies on annotated face images captured using camera/eye-tracker and learns a model to map gaze direction from the images. Most of the earliest methods relied on dedicated feature extraction \cite{rattarom2019model}\cite{yilmaz2016local} and feature selection steps, before performing the regression task \cite{aunsri2022novel}. This in effect makes the models more person-specific and generalization is not attained. At the same time, dedicated feature extraction also makes the computation time-taxing.  With the advancements in deep learning \cite{pathirana2022eye}, convolution neural networks(CNN) based frameworks \cite{cheng2021appearance}\cite{lemley2019convolutional} have become the most used architectures for gaze estimation. Mostly ResNet based CNN frameworks \cite{kanade2021convolutional} have been developed that maps the gaze angles on the eye images. Dual branch CNN \cite{zhu2022complementary} has been another adopted approach for gaze estimation that reconstructs images with the gaze direction in supervision. Sequential models have also been used to learn the variation in gaze direction in subsequent frames, by leveraging LSTM models \cite{chong2020detecting} on the residual features obtained using CNN. Capsule networks are another most promising approach for appearance-based models \cite{bernard2021eye}\cite{mahanama2020gaze} as it emphasizes on learning equivariance instead of finding deeper features. 

Generalization of appearance-based models is solely dependent on the volume of labeled data. Annotation of gaze direction in eye images is a difficult and time-consuming task. Even with the large gaze datasets available in the public domain, there are still issues in developing a domain-adaptive model due to varied background environment and illumination conditions. To deal with these problems, Semi-supervised learning(SSL) is one of the most promising solution. 

SSL performs a pre-training \cite{crawford2019spatially} on the unlabeled data to learn an effective representation of the input. It further fine-tunes the solution using a small labeled dataset. Contrastive learning, a SSL technique, has been predominantly used in different computer vision applications in past few years. Contrastive learning based methods learn semantic embedding representation \cite{chen2020simple} of the input image by pulling the similar images together and pushing the disparate images away \cite{grill2020bootstrap}. This facilitates the models to learn a suitable encoding of the images \cite{zbontar2021barlow} using only the unlabeled data. Consequently, a limited gaze annotations are used to learn a model that can perform the final task of classification, segmentation or prediction. 
Wang \cite{wang2022contrastive} have first introduced the use of contrastive learning method for unsupervised regression learning of eye gaze direction.

In this paper, we present a semi-supervised regression technique to predict the eye gaze direction with two major contributions.
 \begin{itemize}
	\item We have developed a contrastive learning framework that learns an encoder architecture to compute embeddings and further uses the pre-trained embeddings to predict the gaze directions.
	\item We have proposed a new form of contrastive loss to maximize the agreement between similar images that can take care of both invariance and redundancy factors simultaneously.
\end{itemize}

\section{Methodology}
The designed framework is based on SimCLR framework \cite{chen2020simple} that performs the task in two stages. The first stage aims to learn an representation from the given images by maximizing the agreement between the vector representations learned from the images. The final stage uses the embeddings learnt in the pre-training stage and trains a model that performs the task of prediction by minimizing a loss function. 

Mini-batches of batch size $B$ are sampled from the dataset $(X,G_{dir})$, where $X$ represents the images and $G_{dir}$ represents the gaze direction labels for those images. Random data augmentation is performed to create two varied representations of the images in the mini-batches $X_{a1}$ and $X_{a2}$. We have designed an encoder module $\mathcal{E}$  that learns latent space representations $f_1=\mathcal{E}(X_{a1})$ and $f_2=\mathcal{E}(X_{a2})$ from the augmented images. Latent space embeddings learns high-level features from the images. These embeddings are then passed to a projection head $\mathcal{P}$, which is designed as a multi-layer perceptron (MLP) to learn a non-linear projection vector for less complex processing. The learned embeddings are $p_1=\mathcal{P}(f_1)$ and $p_2=\mathcal{P}(f_2)$, which are then compared to minimize the contrastive loss for similar image pairs.

\begin{figure}
	\centering
	\begin{subfigure}{0.8\textwidth}
		\includegraphics[width=0.99\linewidth]{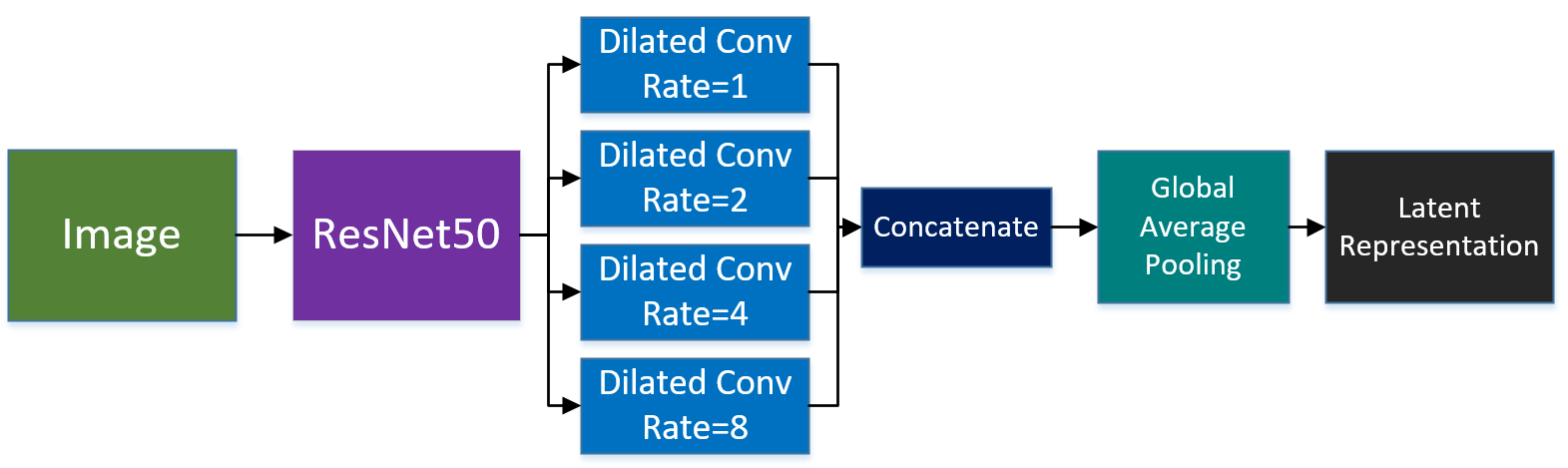}
		\caption{Encoder Architechture}
		\label{encoder}
	\end{subfigure}
	\begin{subfigure}{0.19\textwidth}
		\includegraphics[width=1\linewidth]{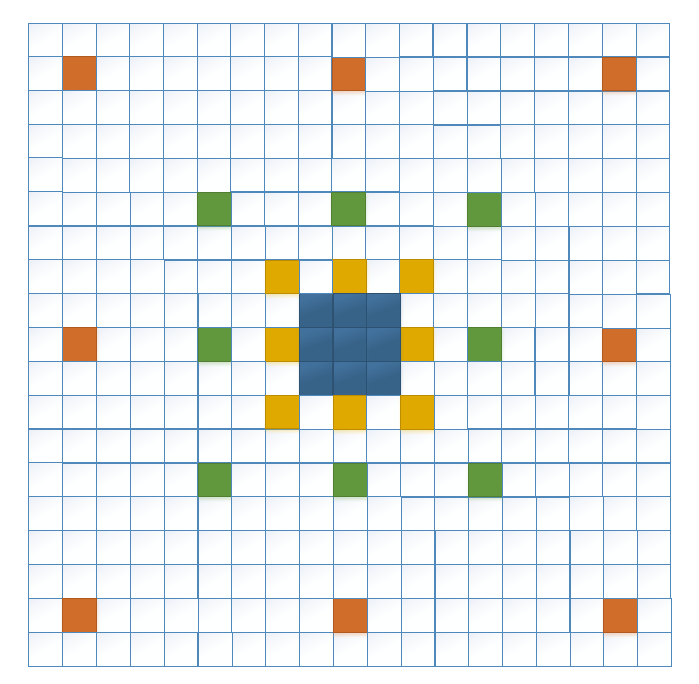}
		\caption{Dilated Convolution for feature mapping}
		\label{Block_Diagram}
	\end{subfigure}
	\caption{}
\end{figure}

The designed encoder can learn local as well as global dependency in feature maps. The architecture of the encoder is shown in Fig.\ref{encoder}. In order to learn global spatial dependency along with local spatial details, larger kernel size needs to be considered. But, larger kernel size increases the computation complexity of the model and can also lead to overfitting. Thus we have used dilated convolution with different dilation rates and local spatial dependency is learnt with each convolution. By using dilated convolution, we were able to attain larger receptive fields without increasing the computation complexity due to larger kernel size by using different dilation rates. Feature map of high spatial resolution is obtained by concatenating the coarse to fine feature maps learned. Convolution filters for different dilation rates have been shown in Fig.\ref{Block_Diagram}, where each colour indicates kernel corresponding to each dilation rate in concatenated feature map.

An updated version of the Normalized temperature scaled cross-entropy (NT-Xent) loss \cite{chen2020simple} has been used to define our contrastive loss. The aim is to minimize the invariance in data as well as the redundancy in feature maps at the same time. The NT-Xent loss aims to maximize the similarity between two positive images and minimize the similarity between a positive and a negative image. We have used the concept of computing cross-correlation matrix as introduced in Barlow Twins \cite{zbontar2021barlow} to analyze the correlation between the positive and negative images. The cross-correlation between the projections $p_1$ and $p_2$ is computed as $C$, where each element basically represents the similarity between each element of the projections. The contrastive loss defined in this paper is displayed in Equation [\ref{contrastive_loss}], where the first term defines the \textbf{NT-Xent loss} that tries to \textbf{minimize the invariance}, whereas the second term \textbf{reduces the redundant} information in the output vector representation. The term $\gamma$ represents loss coefficient factor to consider the redundancy term which has been set to $\gamma=0.01$.  The terms $sim(p_i,p_j)$ defines the cosine similarity between $p_i$ and $p_j$ and is defined by Equation [\ref{cosine_sim}].
\begin{equation} \label{contrastive_loss}
	\mathcal{L}=-\frac{1}{B}\sum_{i,j\in1}^{B} log\frac{exp(sim(p_i,p_j)/\tau)}{\sum_{k=1}^{2B}\mathbbm{1}_{k\neq i}\ exp(sim(p_i,p_k)/\tau)} + \gamma \sum_{i}^{}\sum_{j\neq i}^{}C_{ij}^2
\end{equation}

\begin{equation} \label{cosine_sim}
	sim(p_i,p_j)=\frac{p_i p_j}{\norm{p_i}\norm{p_j}}
\end{equation}

\begin{figure}
	\centering
	\begin{subfigure}{0.49\textwidth}
	\includegraphics[width=1\linewidth]{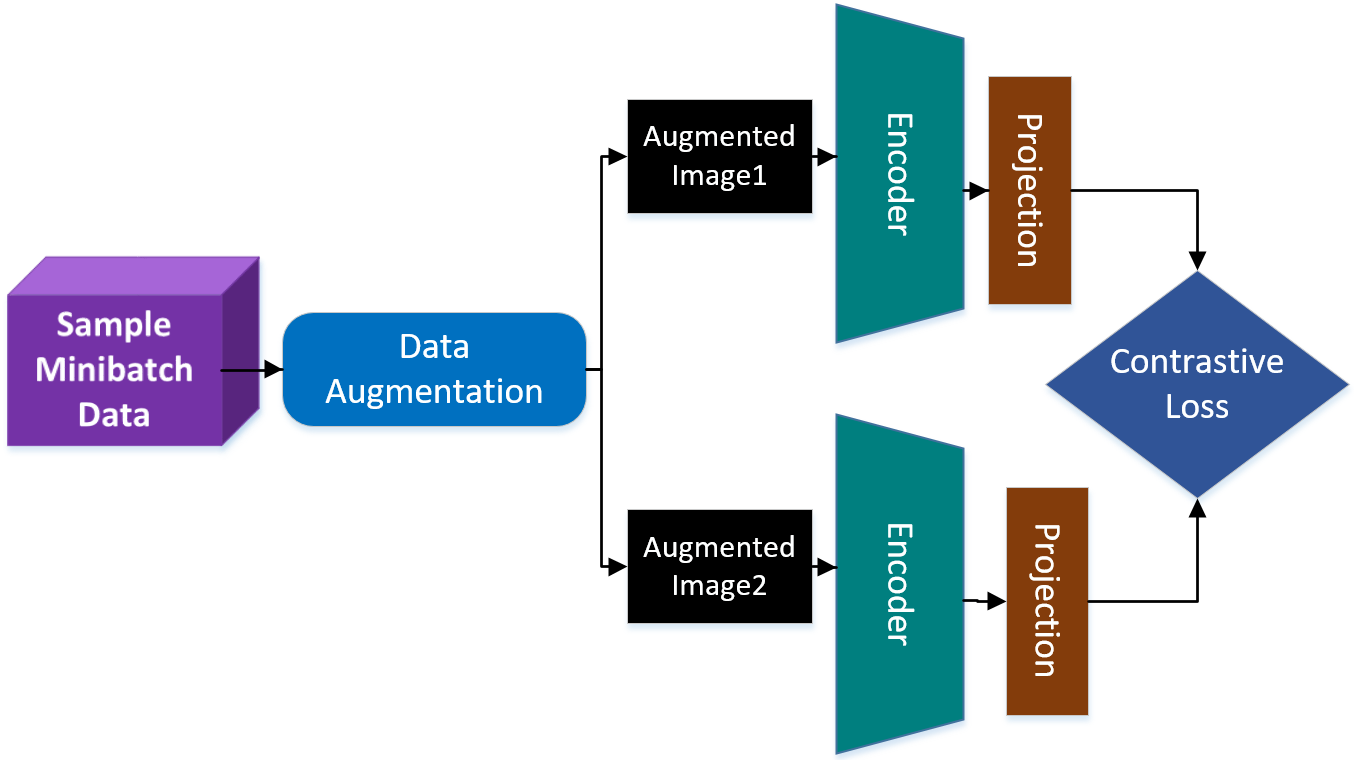}
	\caption{Pre-training Stage}
	\label{pre-training}
\end{subfigure}
\begin{subfigure}{0.5\textwidth}
	\includegraphics[width=1\linewidth]{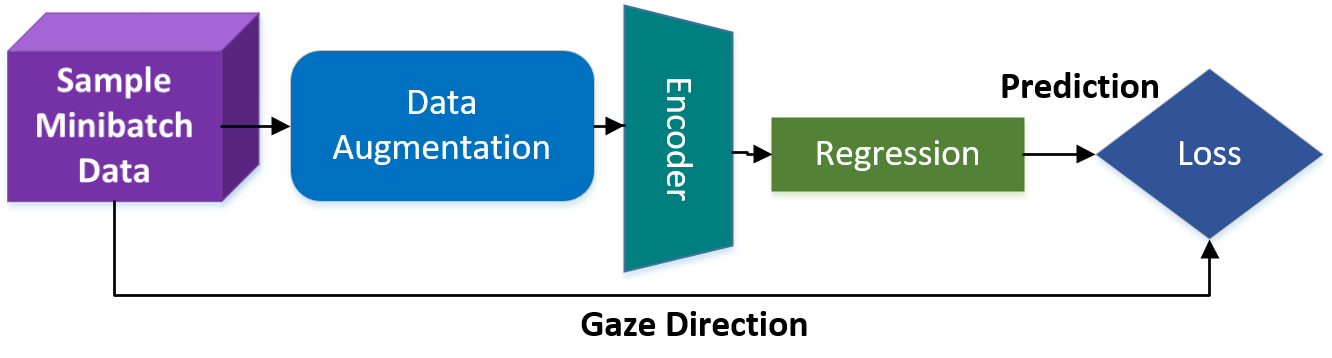}
	\caption{Finetuning Stage}
	\label{regression}
\end{subfigure}
\caption{Contrastive Regression Framework}
\end{figure}

The framework pre-training stage and fine-tuning stage for regression is shown in Fig.\ref{pre-training} and Fig.\ref{regression} respectively. It uses the pre-trained encoder model and determines the values of latent space representations $f_i$. The latent vectors are used as an input to the regression network to predict the values of gaze directions $G_{pred}$. The regression fine-tuning network is optimized by minimizing the huber loss, which is a step-wise amalgamation of mean squared error and mean absolute error to tackle the outlier issue, as defined in Equation [\ref{huber_loss}] where $d=G_{pred}-G_{dir}$. The loss parameter $\delta$, that provides a measure of spread of the deviation error, ensures the loss function to generate large value of loss, by computing the squared value of deviation, for larger deviation values and smaller values of loss, by computing linear mean absolute error, for small deviation values.

\begin{equation} \label{huber_loss}
	L_\delta(d)=
	\begin{cases}
		\frac{d^2}{2}, & \text{if}\ \lvert d\rvert\le\delta \\
		\delta\left(\lvert d\rvert-\frac{\delta}{2}\right), & \text{otherwise}
	\end{cases}
\end{equation}

\section{Results and Discussion}
\subsection{Dataset}
\begin{figure}
	\centering
	\includegraphics[height=1cm,width=0.75\linewidth]{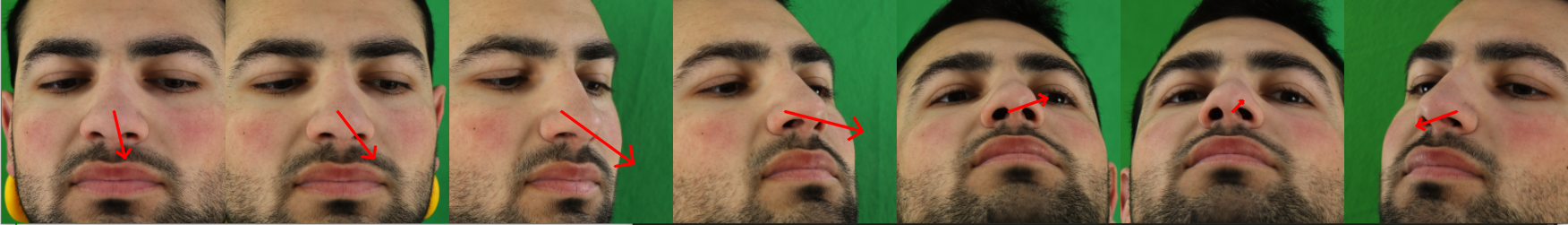}
	\caption{Labeled Face images with Gaze direction}
	\label{xgaze_images}
\end{figure}

\begin{figure}
	\centering
	\begin{subfigure}{0.9\textwidth}
		\includegraphics[height=1cm,width=\linewidth]{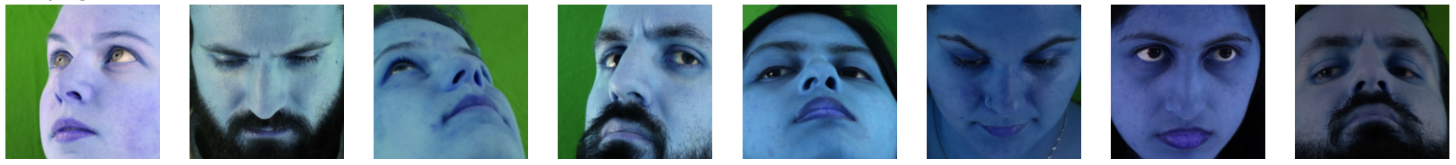}
		\caption{}
	\end{subfigure}
	\begin{subfigure}{0.9\textwidth}
		\includegraphics[height=1cm,width=\linewidth]{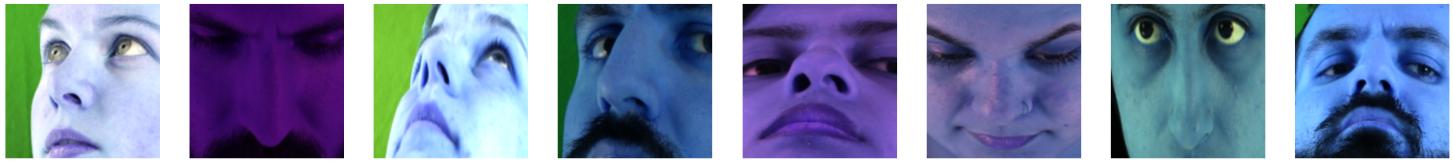}
		\caption{}
	\end{subfigure}
	\caption{Augmented Face images, a. Weak augmentation, b. Strong augmentation}
	\label{xgaze_aug}
\end{figure}

We have used the ETH-XGaze \cite{zhang2020eth} dataset for evaluation of our model. ETH-XGaze is a large data set for gaze estimation that includes images with excellent resolution with consistent label quality, captured from 110 individuals representing a wide range of ages, genders, and ethnicities. The dataset contains around 1,083,492 images of 6000*4000 resolution with 18 cameras placed at different locations to get different views. The dataset covers a large scale of head-pose and gaze directions, which makes it a good choice for developing a generalized solution. Sample images with different gaze directions have been displayed in Fig. \ref{xgaze_images}. 
Images from the train subset of the dataset has been first split into train and validation sets in 80:20 ratio. Train set images, without labels, are used to learn encoder in pre-training stage where validation set images with labels are used further in the fine-tuning stage.
Augmented version of the face images are obtained by applying the following transfer functions such as, horizontal mirror imaging, rescaling, zooming and varying the brightness, contrast, hue and saturation. We have performed augmentation of mini-batches of images by randomly applying any three of the above mentioned operations. Fig. \ref{xgaze_aug} represents few samples of augmented images generated by weak and strong augmentation.

\subsection{Evaluation}
The performance of developed architecture has been evaluated by computing the Mean Angular Error. The performance of our model has been compared with different state of the art architectures and the results have been enlisted in Table [\ref{performance}]. Our model is able to generate significantly better result than SimCLR, which has been used for gaze estimation, and Barlow twins model. Our encoder computes features in an efficient manner by taking local dependencies in consideration. The encoder, when designed with a flatten layer at the output, provides slightly better result. We still prefer the model with global average pooling as displayed in Fig. \ref{encoder}, as it reduces the number of parameters in the architecture. We can see that the model with flatten layer estimates with mean angular error of $2.152$ degrees, whereas the model with reduced parameters generates mean angular error of $3.212$ degrees.
\begin{table}[h!]
	\begin{center}
		\caption{Mean Angular Error in degrees for gaze estimation}
		\label{performance}
		\begin{tabular}{l|c} 
			\hline
			{Method} & {Mean Angular Gaze Error} \\
			\hline
			SimCLR & 9.175\textdegree \\
			Barlow Twins & 9.390\textdegree \\
			SimCLR with Deeplab Encoder & 2.246\textdegree \\
		    \textbf{Ours} & \textbf{2.152\textdegree} \\
			Ours (with reduced parameters) & 3.212\textdegree \\
			\hline
		\end{tabular}
	\end{center}
\end{table}

We have evaluated performance of models with other state of the art loss functions used in contrastive training for gaze estimation problem. The simultaneous optimization of invariance and redundancy provides us the upper hand in performance as shown in Table [\ref{ablation_loss}]. 
\begin{table}[h!]
	\begin{center}
		\caption{Ablation study on using different Loss}
		\label{ablation_loss}
		\begin{tabular}{l|c} 
			\hline
			{Method} & {Mean Angular Gaze Error } \\
			\hline
			NT-Xent loss & 4.606\textdegree \\
			Barlow Twins loss & 4.3361\textdegree \\
			Our Contrastive loss & 3.212\textdegree \\
			\hline
		\end{tabular}
	\end{center}
\end{table} 
      
\subsection{Ablation Study}
In order to understand the role of second term in loss function defined in Equation [\ref{contrastive_loss}], which aims to minimize the redundancy, we have used a loss coefficient factor $\gamma$. By varying the value of $\gamma$, the performance of our model is evaluated. In Table [\ref{ablation_parameter}], we can see that the performance improves as we increase the value of loss coefficient upto a certain value. The coefficient $\gamma$ has been set as 0.1 for evaluation of our model.
\begin{table}[h!]
	\begin{center}
		\caption{Ablation study on loss coefficient parameter}
		\label{ablation_parameter}
		\begin{tabular}{l|c} 
			\hline
			{Method} & {Mean Angular Gaze Error } \\
			\hline
			$\gamma=0.005$ & 4.755\textdegree \\
			$\gamma=0.01$ & 4.549\textdegree \\
			$\gamma=0.1$ & 3.212\textdegree \\
			\hline
		\end{tabular}
	\end{center}
\end{table}

\section{Conclusion}
This work has enlightened on semi-supervised learning for eye gaze prediction tasks by developing contrastive learning framework. The work has proposed a new form of contrastive loss to optimize the similarity agreement between augmented and captured face images considering two parameters, viz., invariant transformation and redundancy among images, to predict gaze direction. The proposed model has been tested on ETH-XGaze dataset. The evaluation of the proposed model has been done in terms of mean angular error. The model has outperformed different state of the art techniques such as SimCLR, Barlow twins as shown in the previous sections. In future we will experiment across different datasets and find a more generalized solution for cross-datasets scenarios.

\subsubsection{Acknowledgements}
Authors would like to acknowledge CSIR-Central Electronics Engineering Research Institute (CSIR-CEERI) for providing facilities and CSIR-AITS mission for providing fund to conduct this research work.

%
%
\bibliographystyle{splncs}
\bibliography{refbank_gaze}	

\begin{thebibliography}{10}

\bibitem{konrad2020gaze}
Konrad, R., Angelopoulos, A., Wetzstein, G.:
\newblock Gaze-contingent ocular parallax rendering for virtual reality.
\newblock ACM Transactions on Graphics (TOG) \textbf{39}(2) (2020)  1--12

\bibitem{gerber2020self}
Gerber, M.A., Schroeter, R., Xiaomeng, L., Elhenawy, M.:
\newblock Self-interruptions of non-driving related tasks in automated
  vehicles: Mobile vs head-up display.
\newblock In: Proceedings of the 2020 CHI Conference on Human Factors in
  Computing Systems. (2020)  1--9

\bibitem{ferrier2022measuring}
Ferrier-Barbut, E., Gauthier, P., Luengo, V., Canlorbe, G., Vitrani, M.A.:
\newblock Measuring the quality of learning in a human--robot collaboration: A
  study of laparoscopic surgery.
\newblock ACM Transactions on Human-Robot Interaction (THRI) \textbf{11}(3)
  (2022)  1--20

\bibitem{rattarom2019model}
Rattarom, S., Uttama, S., Aunsri, N.:
\newblock Model construction and validation in low-cost interpolation-based
  gaze tracking system.
\newblock Engineering Letters \textbf{27}(1) (2019)

\bibitem{yilmaz2016local}
Yilmaz, C.M., Kose, C.:
\newblock Local binary pattern histogram features for on-screen eye-gaze
  direction estimation and a comparison of appearance based methods.
\newblock In: 2016 39th International Conference on Telecommunications and
  Signal Processing (TSP), IEEE (2016)  693--696

\bibitem{aunsri2022novel}
Aunsri, N., Rattarom, S.:
\newblock Novel eye-based features for head pose-free gaze estimation with web
  camera: New model and low-cost device.
\newblock Ain Shams Engineering Journal \textbf{13}(5) (2022)  101731

\bibitem{pathirana2022eye}
Pathirana, P., Senarath, S., Meedeniya, D., Jayarathna, S.:
\newblock Eye gaze estimation: A survey on deep learning-based approaches.
\newblock Expert Systems with Applications \textbf{199} (2022)  116894

\bibitem{cheng2021appearance}
Cheng, Y., Wang, H., Bao, Y., Lu, F.:
\newblock Appearance-based gaze estimation with deep learning: A review and
  benchmark.
\newblock arXiv preprint arXiv:2104.12668 (2021)

\bibitem{lemley2019convolutional}
Lemley, J., Kar, A., Drimbarean, A., Corcoran, P.:
\newblock Convolutional neural network implementation for eye-gaze estimation
  on low-quality consumer imaging systems.
\newblock IEEE Transactions on Consumer Electronics \textbf{65}(2) (2019)
  179--187

\bibitem{kanade2021convolutional}
Kanade, P., David, F., Kanade, S.:
\newblock Convolutional neural networks (cnn) based eye-gaze tracking system
  using machine learning algorithm.
\newblock European Journal of Electrical Engineering and Computer Science
  \textbf{5}(2) (2021)  36--40

\bibitem{zhu2022complementary}
Zhu, Z., Zhang, D., Chi, C., Li, M., Lee, D.J.:
\newblock A complementary dual-branch network for appearance-based gaze
  estimation from low-resolution facial image.
\newblock IEEE Transactions on Cognitive and Developmental Systems (2022)

\bibitem{chong2020detecting}
Chong, E., Wang, Y., Ruiz, N., Rehg, J.M.:
\newblock Detecting attended visual targets in video.
\newblock In: Proceedings of the IEEE/CVF conference on computer vision and
  pattern recognition. (2020)  5396--5406

\bibitem{bernard2021eye}
Bernard, V., Wannous, H., Vandeborre, J.P.:
\newblock Eye-gaze estimation using a deep capsule-based regression network.
\newblock In: 2021 International Conference on Content-Based Multimedia
  Indexing (CBMI), IEEE (2021)  1--6

\bibitem{mahanama2020gaze}
Mahanama, B., Jayawardana, Y., Jayarathna, S.:
\newblock Gaze-net: Appearance-based gaze estimation using capsule networks.
\newblock In: Proceedings of the 11th augmented human international conference.
  (2020)  1--4

\bibitem{crawford2019spatially}
Crawford, E., Pineau, J.:
\newblock Spatially invariant unsupervised object detection with convolutional
  neural networks.
\newblock In: Proceedings of the AAAI Conference on Artificial Intelligence.
  Volume~33. (2019)  3412--3420

\bibitem{chen2020simple}
Chen, T., Kornblith, S., Norouzi, M., Hinton, G.:
\newblock A simple framework for contrastive learning of visual
  representations.
\newblock In: International conference on machine learning, PMLR (2020)
  1597--1607

\bibitem{grill2020bootstrap}
Grill, J.B., Strub, F., Altch{\'e}, F., Tallec, C., Richemond, P., Buchatskaya,
  E., Doersch, C., Avila~Pires, B., Guo, Z., Gheshlaghi~Azar, M.,  et~al.:
\newblock Bootstrap your own latent-a new approach to self-supervised learning.
\newblock Advances in neural information processing systems \textbf{33} (2020)
  21271--21284

\bibitem{zbontar2021barlow}
Zbontar, J., Jing, L., Misra, I., LeCun, Y., Deny, S.:
\newblock Barlow twins: Self-supervised learning via redundancy reduction.
\newblock In: International Conference on Machine Learning, PMLR (2021)
  12310--12320

\bibitem{wang2022contrastive}
Wang, Y., Jiang, Y., Li, J., Ni, B., Dai, W., Li, C., Xiong, H., Li, T.:
\newblock Contrastive regression for domain adaptation on gaze estimation.
\newblock In: Proceedings of the IEEE/CVF Conference on Computer Vision and
  Pattern Recognition. (2022)  19376--19385

\bibitem{zhang2020eth}
Zhang, X., Park, S., Beeler, T., Bradley, D., Tang, S., Hilliges, O.:
\newblock Eth-xgaze: A large scale dataset for gaze estimation under extreme
  head pose and gaze variation.
\newblock In: Computer Vision--ECCV 2020: 16th European Conference, Glasgow,
  UK, August 23--28, 2020, Proceedings, Part V 16, Springer (2020)  365--381

\end{thebibliography}

\end{document}